\pdfoutput=1

\documentclass[10pt,twocolumn,letterpaper]{article}
\usepackage{iccv}

\usepackage[accsupp]{axessibility} 
\usepackage{times}
\usepackage{epsfig}
\usepackage{graphicx}
\usepackage{amsmath}
\usepackage{amssymb}
\usepackage{dsfont}
\usepackage{pifont}
\usepackage{enumitem}
\usepackage{xcolor}
\newcommand{\xmark}{\ding{55}}%
\usepackage{amsfonts}
\usepackage{dsfont}

\usepackage{soul}
\usepackage{subfigure}
\usepackage{amsfonts}
\usepackage{amsthm}
\usepackage{booktabs}
\usepackage{arydshln}

\setlist[itemize]{leftmargin=4.mm}

\usepackage[pagebackref=true,breaklinks=true,colorlinks,bookmarks=false]{hyperref}

\usepackage{color, colortbl}
\definecolor{LightCyan}{rgb}{0.88,1,1}
\newcommand{\comment}[1]{}

\iccvfinalcopy 

\def\iccvPaperID{7999}

\ificcvfinal\fi

\usepackage{authblk}
\begin{document}

\title{In Defense of Scene Graphs for Image Captioning}


\author[1]{Kien Nguyen\thanks{Authors have equal contributions}}
\newcommand\CoAuthorMark{\footnotemark[\arabic{footnote}]} 
\author[2]{Subarna Tripathi\protect\CoAuthorMark}
\author[1]{Bang Du}
\author[3]{Tanaya Guha}
\author[1]{Truong Q. Nguyen}
\affil[1]{University of California San Diego, USA}
\affil[2]{Intel Labs, USA}
\affil[3]{University of Warwick, UK}
\renewcommand\Authands{, and }

\maketitle

\label{abs}
\begin{abstract}
The mainstream image captioning models rely on 
Convolutional Neural Network (CNN) image features 
to generate captions via recurrent models. Recently, image scene graphs have been used to augment captioning models so as to leverage their structural semantics, such as object entities, relationships and attributes. 
Several studies have noted that the naive use of scene graphs from a black-box scene graph generator harms image captioning performance and that scene graph-based captioning models have to incur the overhead of explicit use of image features to generate decent captions. Addressing these challenges, we propose \textbf{SG2Caps}, a framework that utilizes only the scene graph labels for competitive image captioning performance. The basic idea is to close the semantic gap between the two scene graphs - one derived from the input image and the other from its caption. In order to achieve this, we leverage the spatial location of objects and the Human-Object-Interaction (HOI) labels as an additional HOI graph. SG2Caps outperforms existing scene graph-only captioning models by a large margin, indicating scene graphs as a promising representation for image captioning. Direct utilization of scene graph labels avoids expensive graph convolutions over high-dimensional CNN features resulting in $49\%$ fewer trainable parameters. Our code is available at: \url{https://github.com/Kien085/SG2Caps}
\end{abstract}

\section{Introduction}
\label{sec:intro}
%

The mainstream image captioning models rely on convolutional image features and/or attention to salient regions and objects to generate captions via recurrent models \cite{ShowTell_vinyals2014neural,Anderson_2018_CVPR}. Recently, \textbf{\emph{scene graph}} representations of images have been used to augment captioning models so as to leverage their structural semantics, such as object entities, relationships and attributes ~\cite{Vis_rel_Yao_2018_ECCV,SC_autoencoding_Yang_2019_CVPR,unpaired_caption_iccv19}. The literature however has mixed opinion about the usefulness of scene graphs in captioning. Few works have reported improvements in caption generation using scene graphs \cite{lanterns19,SC_autoencoding_Yang_2019_CVPR}, while several others have highlighted that scene graphs alone yield poor captioning results and can even harm captioning performance \cite{tmm19,AACL-IJCNLP2020}. In this paper, we identify the challenges in effective utilization of scene graphs for image captioning, and subsequently investigate how to best harness them for this task. 

\begin{figure*}[t] 
\centering
    \includegraphics[width=1\linewidth]{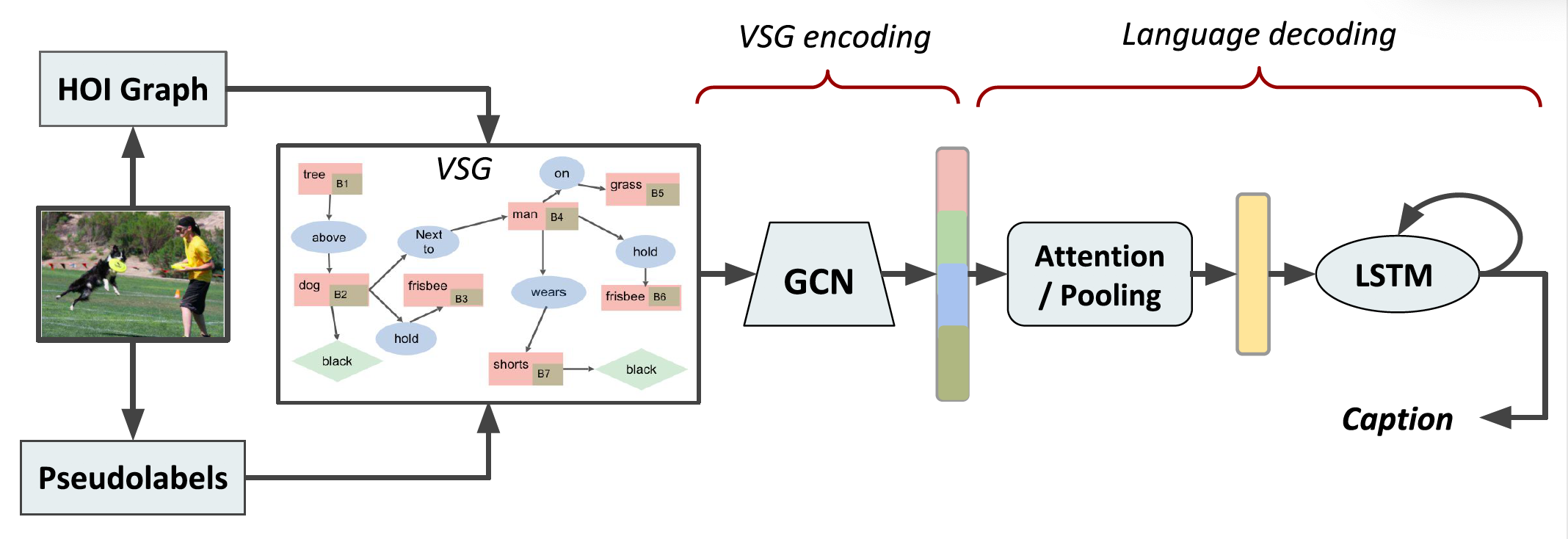}
\caption{{\bf SG2Caps} first creates Visual Scene Graphs (VSG) by combining (1) pseudolabel - output of a black-box VSG generator, and (2) HOI graph from an HOI inference model. Each object node of the VSG has a bounding box label. Object nodes, relations, attributes are color-coded in red, blue, green respectively. The output of VSG encoding is the input for the LSTM-based decoder for the caption generation.
}
\label{fig:SG2Caps}
\end{figure*}

Scene graph representation consisting of nodes and edges can be derived from either (i) images where the nodes correspond to the objects present in the scene, termed as \emph{Visual Scene Graphs \textbf{(VSG)}}, or (ii) from a caption where nouns and verbs take on the roles of nodes and edges in a rule-based semantic parsing, termed as \emph{Textual Scene Graphs \textbf{(TSG)}}. 
The literature of scene graph generation and scene graphs for image captioning primarily refers to the VSG representation. 

To be able to leverage scene graphs for captioning, we need paired VSG-caption annotations. This is currently unavailable. Hence, methods requiring explicit scene graphs end up training the VSG generator and the caption generator on disparate datasets \cite{SC_autoencoding_Yang_2019_CVPR,unpaired_caption_iccv19,AACL-IJCNLP2020}. The current practice is to train VSG generators on the Visual Genome (VG) dataset, 
train TSG to caption generation on COCO-captions dataset, and finally transform the VG-trained VSGs to captions utilizing the later. 
We note two issues with this approach: \\
$\bullet$ 
The VG-trained VSGs are highly biased towards certain types of relationships (e.g., \emph{has, on}); the relationship distribution is significantly long-tailed, and even the top-performing VSG generators fail to learn meaningful relationships accurately ~\cite{survey2020}. This results in noisy VSGs, which in turn degrades the quality of captions ~\cite{AACL-IJCNLP2020}. 
\begin{figure}[t] 
\centering
    \subfigure[An image and the TSG generated from its caption]
    {\includegraphics[width=1\linewidth]{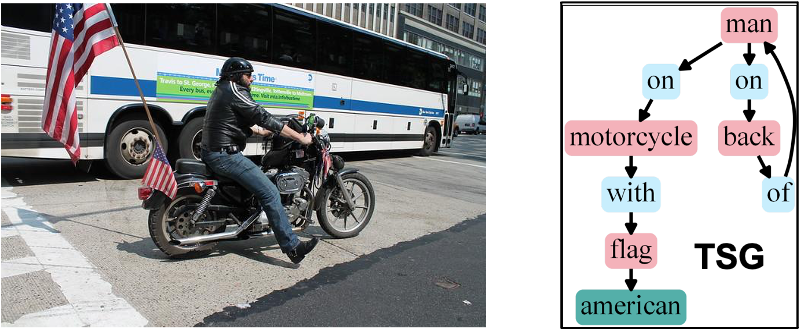} 
    }
    \subfigure[VSG containing all detected objects as nodes ]
    {\includegraphics[width=1\linewidth]{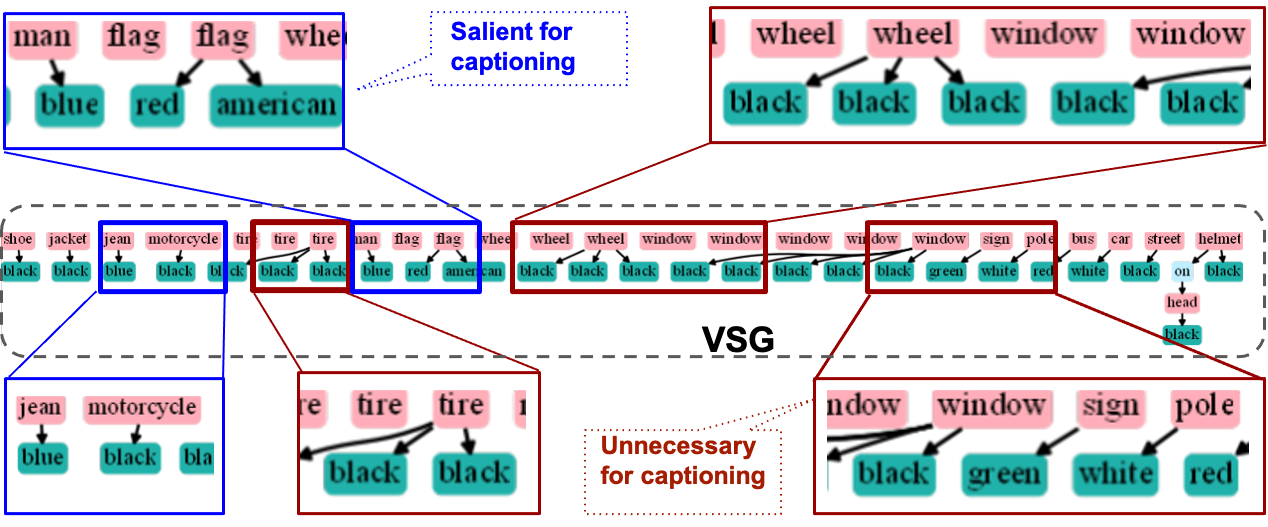} 
    }
    \caption{Characterization of TSG and VSG. While TSG only contains salient contents such as \emph{man, motorcycle, flag} for natural language description, VSG includes unnecessary details such as \emph{wheel, tire, window, sign, pole}. Objects, attributes, edges are shown in \emph{pink,green,blue} respectively. (Best viewed in color)}
\label{fig:sg_viz}
\end{figure}

\noindent 
$\bullet$ 
There is an assumption in the existing approaches that TSGs and VSGs are compatible. But, \textit{are VSGs and TSGs actually compatible?}  
TSGs, when used as inputs, can generate excellent captions \cite{SC_autoencoding_Yang_2019_CVPR}. 
However, the problem arises when parameters trained for TSGs are used for VSG inputs, assuming direct compatibility. TSGs, being generated from captions, do not include every object seen in the image or all their pairwise relationships - the very information VSGs are designed to extract (see Fig.~\ref{fig:sg_viz}).
In other words, VSGs are exhaustive while TSGs focus only on the salient objects and relationships. 
Thus natural language inductive bias does not translate automatically from TSG models to VSG models. We argue that this is the main reason why previous efforts to exploit VSGs for captioning did not achieve desired results. 
\par 
To mitigate the above issues, we explore several novel ways to enhance VSGs in the context of captioning:\\
(i) \textbf{Human-Object Interaction (HOI) information:} Humans tend to describe visual scenes involving humans by focusing on the human-object interactions at the exclusion of other details.
If HOI information is extracted from an image, it can provide an effective way to highlight the `salient' parts in its VSG, thereby bringing it closer to its corresponding TSG. Hence, we propose to harness pre-trained HOI inferences as \emph{partial} VSGs, where all detected objects (not limited to humans) in a scene form the graph nodes and the HOI information augment a few relevant nodes with appropriate relationship and attributes.\\
(ii) \textbf{VSG grounding:} A unique aspect of VSG is that each of the node in an VSG is grounded, i.e., has a one-to-one association with the object bounding boxes in the image. This spatial information can be used to capture the relationship between objects. 
It is well known from the scene graph generation literature 
that the inter-object relationship classification performance greatly benefits from ground-truth bounding box locations \cite{VCTree_Tang_2019_CVPR,tang2020unbiased}. Despite this evidence, no VSG-based captioning model has yet used the spatial information of the nodes. We show that such information can significantly improve captioning performance. 
 

\par
This paper investigates how to best leverage VSGs for caption generation, if at all. To this end, we develop a new image captioning model, termed \textbf{SG2Caps}, that  utilizes the VSGs \emph{alone} for caption generation 
(see Fig.~\ref{fig:SG2Caps} for the main idea). In contrast to the existing work, we do not use any image or object-level visual features; yet, we achieve competitive caption generation performance by exploiting the HOI information and VSG grounding. 
Directly utilizing the scene graph labels avoids expensive graph convolutions over high-dimensional CNN features, we show that it is still effective for caption generation via capturing visual relationships.  
This also results in $\mathbf{49\%}$ reduction in the number of trainable parameters comparing with the methods that require processing of both visual features and scene graphs. Researchers have shown that image captioning algorithms produce less accurate results due to inherent dataset biases and the unavailability of high quality human annotations. ~\cite{Burns2018WomenAS}. Our \textbf{SG2Caps} model leverages additional annotations from other sources beyond image caption datasets thereby reducing NLP bias.
\par Our contributions are summarized below.
\begin{itemize}
\setlength\itemsep{0mm}
\item 
We show that competitive captioning performance can be achieved on the COCO-captions dataset using VSG alone, \emph{without} any visual features. 
\item We experimentally show that VSGs and TSGs are not compatible with each other in the context of caption generation, 
and we propose to improve the learnable transformation directly from VSG to caption.
\item We propose a new captioning model, \textbf{SG2Caps}, that utilizes spatial locations of VSG nodes and HOI information to generate better caption from VSGs. While node locations help to identify the meaningful relationships among objects (results in 5 point gain in CIDEr score), HOI captures the essence of natural language communication (results in 7 point gain in CIDEr score). Thus they help to close semantic gap between TSGs and VSGs for the purpose of image captioning.
\item  
We also devise an extremely light-weight visual feature fusion strategy for the SG2Caps framework.  
The low-dimensional global visual features added as a summary node in the VSG, 
boosts image captioning performance atop blackbox scene-graph only model. 
Our \textbf{SG2Caps} model performs competitively against existing captioning models that operate on high-dimensional region level visual features.


\end{itemize}

\section{Related work}
\label{sec:related}
\textbf{Image captioning:}
Mainstream image captioning models ~\cite{ShowTell_vinyals2014neural,LongTerm_Donahue_2015_CVPR} directly feed convolutional image features to a recurrent network to generate natural language. 
The \emph{top-down} approaches in such image captioning models rely on attention-based deep models ~\cite{Lu2017Adaptive,rennie_self-critical_2017,NIPS2016_9996535e,show_attend_tell_15} 
where a partially-completed caption is utilized as the `context'. An attention, based on the context, is then applied to the output of one or more layers of a CNN. These approaches can predict attention weights without any prior knowledge of objects or salient image regions.
%
Later, \emph{bottom-up} approaches ~\cite{Anderson_2018_CVPR} enabled attention to be computed at the object-level using object detectors. Such object-level attention mechanism is the state-of-the-art in many vision-language tasks including image captioning and visual question answering. 

\par
\textbf{VSG in image captioning}: 
A number of works \cite{PCPL_YanSJHJC020,Gu_2019_CVPR,VCTree_Tang_2019_CVPR,xu2017scenegraph,graph_RCNN_Yang_2018_ECCV,FactorizableNet_Li_2018_ECCV,neural_motifs_2017} 
devised approaches that strive to perform on VSG generation tasks on the benchmark VG dataset \cite{visual_genome16}.
A few recent works ~\cite{Vis_rel_Yao_2018_ECCV,SC_autoencoding_Yang_2019_CVPR,unpaired_caption_iccv19,lanterns19,jvcir18,tmm19}
have introduced the use of VSG (in addition to the visual features) with the hope that encoding of objects attributes and relationships would improve image captioning. Some of the works used implicit scene graph representation ~\cite{Vis_rel_Yao_2018_ECCV,Gao2018_icmlc}, while others explored an explicit representation 
of 
relations and attributes  ~\cite{SC_autoencoding_Yang_2019_CVPR,lanterns19,AACL-IJCNLP2020,visrelprior_2019}. 
The explicit scene graph approaches integrate VSG features with CNN features from image or objects. Such explicit scene graph approaches use a scene graph generator as a blackbox. 
For example, Wang \textit{et al.} \cite{lanterns19} utilized FactorizableNet \cite{FactorizableNet_Li_2018_ECCV}, some 
~\cite{SC_autoencoding_Yang_2019_CVPR,unpaired_caption_iccv19,zhong2020comprehensive} used MotifNet ~\cite{neural_motifs_2017} and others ~\cite{tmm19,AACL-IJCNLP2020} utilized Iterative Message Passing ~\cite{xu2017scenegraph} as their blackbox scene graph generator. 
\par
Researchers found that VSG \emph{alone} yields poor captioning results. The literature so far has mixed opinions about the usefulness of scene graphs. While Wang \textit{et al.} \cite{lanterns19} observed improvement in caption generation using VSG, Li and Jiang \cite{tmm19} did not find VSGs useful. Recently, an in-depth study concluded that it is the noise in VSG (often the relations) that harm the image captioning performance \cite{AACL-IJCNLP2020}. 

In contrast to the existing VSG-augmented captioning models, our model \textbf{SG2Caps} does \emph{not} use object-level CNN features as inputs but utilizes the VSG labels primarily. We are aware of one work that used VSG labels as the only input in captioning and observed that VSG labels alone lead to unsatisfactory results \cite{lanterns19}. Another recent method \cite{zhong2020comprehensive} was designed to utilize both the scene graphs and the region-level visual features.
Our model significantly differs from them in several ways: (1) Our model incorporates novel techniques to make VSGs compatible for captioning; (2) we leverage HOI information to further improve captioning, and (3) our model can handle variable number of nodes in the VSGs. Our model achieves competitive image captioning performances with VSG labels alone. We also outperform the scene graph-only mode adapted from the work of Zhong \textit{et al.} ~\cite{zhong2020comprehensive} that leverages a richer object class vocabulary (1600 vs 150) and external word embeddings. 

\section{Proposed model}
\label{sec:method}
Our model, SG2Caps, consists of a VSG generator, VSG encoder and a language decoder (Fig.~\ref{fig:SG2Caps}). Given an image $I$, the VSG generator constructs a VSG $\mathcal{G}$ that is particularly suited for caption generation. The VSG consists of objects, their attributes, spatial locations and inter-object relationships. The VSG encoder then generates context-aware embeddings, which are input to the language decoder. The language decoder consists of a long-short-term-memory network (LSTM) 
attention followed by an LSTM-based language model that generates the captions. 

\par
\subsection{VSG generator}
Our VSG generator has the following two components.

\indent \textbf{VG150 Pseudolabel:} 
Off-the-shelf VSG generators provide object classes as the node labels and pairwise relationship. 
We learn our own attribute classifier and train a VSG generator on VG150 using MotifNet \cite{neural_motifs_2017}. This pretrained VSG generator is applied on the COCO images to create scene graph pseudolabels so as to create visual scene graphs with nodes, their attributes, their locations and pairwise relationships. The objects, attributes and relationships correspond to the vocabulary of VG150 consisting of 150 object classes, 50 relations and 203 attributes. 
Please note that COCO does not provide scene graph annotations. Also note that pseudolabel path is not bound to the specific one we used in this paper, it can be any black-box visual scene graph generator. 

\textbf{Partial COCO scene graph:} 
In parallel, we use an object detector pretrained on COCO images to create a list of COCO objects that serve as nodes of another graph, termed as HOI graph.  
Then we use an HOI inference to fill up only a few attributes and relationship edges involving only `person' category. We call it a partial scene graph since it has a limited relationship information with mostly disconnected nodes but all in COCO vocabulary. 
Images where \emph{person} category objects are not detected, the partial VSGs do not contain any HOI augmentation, it consists of only the list of nodes 
created from detected object instances of other COCO categories. 
\subsection{VSG encoder} 
We use the union of pseudolabels and HOI graphs as the VSG of an image. A VSG is a tuple $\mathcal{G} = (\mathcal{N},\mathcal{E})$, where $\mathcal{N}$ and $\mathcal{E}$ are the sets of nodes and edges. 
In our formulation, there are four kinds of nodes in $\mathcal{N}$: object nodes $o$, attribute nodes $a$, 
bounding box nodes $b$ and relationship nodes $r$. We denote $o_i$ as the $i$-th object, $r_{ij}$ as the relationship between $o_i$ and $o_j$, 
$b_{i}$ as the bounding box coordinates of $o_i$
and $a_{i,l}$ as the $l$-th attribute of object $o_i$. 
Each node in $\mathcal{N}$ is represented by a $d$-dimensional vector pertaining to the node type \textit{i.e.}, $\mathbf{e}_o$, $\mathbf{e}_a$, $\mathbf{e}_b$ and $\mathbf{e}_r$. 
The edges in $\mathcal{E}$ are defined as follows:
\begin{itemize}
\item If an object $o_i$ has an attribute $a_{i,l}$ there is a directed edge from $o_i$ to $a_{i,l}$.
\item There is a directed edge from $o_i$ to its bounding box $b_{i}$.
\item If there exists a relationship triplet $<o_i - r_{ij} - o_j>$, two directed edges from $o_i$ to $r_{ij}$ and from $r_{ij}$ to $o_j$ are constructed.
\end{itemize}
Next, we transform the original node embeddings $\mathbf{e}_o$, $\mathbf{e}_a$, $\mathbf{e}_b$,
$\mathbf{e}_r$ to a new set of context-aware embeddings $\mathcal{X} = \{ \mathbf{x}_{r_{ij}}, \mathbf{x}_{o_{i}}, \mathbf{x}_{a_{i}},\mathbf{x}_{b_{i}}\}$, where
 $\mathbf{x}_{r_{ij}}$ is the relationship embedding for the relationship node $r_{ij}$, $\mathbf{x}_{o_{i}}$, $\mathbf{x}_{a_{i}}$ and $\mathbf{x}_{b_{i}}$ are the object embedding, attribute embedding and bounding box embedding for the object node $o_{i}$.
We use five spatial graph convolution functions $g_r$, $g_a$, $g_b$, $g_s$, and $g_o$ to generate the above embeddings. 
All of them have same structure with independent parameters: a fully-connected layer, followed by a ReLU.

\noindent\textbf{Relationship embedding  $\mathbf{x}_{r_{ij}}$: } Given a relationship triplet $<o_i - r_{ij} - o_j>$ in $\mathcal{G}$, $\mathbf{x}_{r_{ij}}$ is defined in the context of the subject ($o_i$), object ($o_j$) and predicate ($r_{ij}$) 
together as follows: 
\begin{equation}
    \mathbf{x}_{r_{ij}} = g_r(\mathbf{e}_{o_{i}},\mathbf{e}_{r_{ij}},\mathbf{e}_{o_{j}})
\end{equation}
\noindent
\textbf{Attribute embedding $\mathbf{x}_{a_{i}}$: } For an object node $o_{i}$ with its attributes $a_{i,1:N_{a_i}}$ in $\mathcal{G}$, the embedding $\mathbf{x}_{a_{i}}$ is given by 
\begin{equation}
    \mathbf{x}_{a_{i}} = \frac{1}{N_{a_i}}\sum_{l=1}^{Na_i} g_a(\mathbf{e}_{o_{i}}, \mathbf{e}_{a_{i,l}})
\end{equation}
where $N_{a_i}$ is the number of attributes for $o_i$. Here the context of an object with all its attributes are incorporated.\\
\noindent
\textbf{Bounding box embedding $\mathbf{x}_{b_{i}}$:} Given $o_{i}$ with its bounding box $b_{i}$, $\mathbf{x}_{b_{i}}$ is defined as: 
\begin{equation}
    \mathbf{x}_{b_{i}} = g_b(\mathbf{e}_{o_{i}}, \mathbf{e}_{b_{i}})
\end{equation}
\noindent
\textbf{Object embedding $\mathbf{x}_{o_{i}}$:} An object node $o_{i}$ plays different roles based on the edge directions, \textit{i.e.}, whether $o_{i}$ acts as a subject or the object in a triplet. Following past work ~\cite{SC_autoencoding_Yang_2019_CVPR}, our object embedding takes the entire triplet into consideration. We define $\mathbf{x}_{o_i}$ as follows:
\begin{equation}
\begin{split}
    \mathbf{x}_{o_{i}} = \frac{1}{N_{r_i}} [ \sum_{o_j \in sbj(o_i)}g_s(\mathbf{e}_{o_{i}},\mathbf{e}_{o_{j}}, \mathbf{e}_{r_{ij}}) \\
    + \sum_{o_k \in obj(o_i)}g_o(\mathbf{e}_{o_{k}},\mathbf{e}_{o_{i}}, \mathbf{e}_{r_{ki}})
    ]
\end{split}
\end{equation}
where $N_{r_i} = |sbj(i)| + |obj(i)|$ is the number of relationship triplets involving $o_i$. Each node $o_j \in sbj(o_i)$ acts as an \emph{object} while $o_i$ acts as a \emph{subject}. 

Note that our model differs from SGAE in the ways we learn parameters for $\mathbf{x}_{b_{i}}$ and fuse different output embeddings. We combine $\mathbf{x}_{o_{i}}$, $\mathbf{x}_{b_{i}}$, $\mathbf{x}_{a_{i}}$ with a sum operation at node-level to form $\mathbf{\mathcal{X}}$ before feeding them to the attention LSTM, whereas SGAE simply concatenated them. 

\par
\textbf{Fusing visual features}
We employ a simple strategy to augment the scene graph features optionally with visual features. We add a summary projection node in the output graph embedding. The summary node is generated using global pooling of the image-level P6 features, followed by a projection layer that takes the 256-dimensional summary node feature to 128-dimensional vector, followed by a ReLU non-linearity. The projected summary node dimension is compatible with the GCN embedding size. This projected summary node feature is concatenated with the $\mathbf{\mathcal{X}}$ vector generated as mentioned above. 
The above feature fusion strategy based on summary node also differs from others ~\cite{SC_autoencoding_Yang_2019_CVPR, zhong2020comprehensive} which fuse features at individual node level. 
This feature fusion has a negligible parameter overhead (256$\times$128 additional parameters) on top of SG2Caps. 
\subsection{Language Model}
Given the VSG $\mathcal{G}$ of an image $I$ , we want to generate a natural language sentence $w_{1:T} = {w_1, w_2, ..., w_T }$ that describes the image. In SG2Caps, we follow a 2-layer encoder-decoder LSTM architecture for this part. The encoder, an attention LSTM, takes the VSG encoding $\mathcal{X}$ as input. This is in contrast to the popular captioning models where CNN features are used. The decoder is an LSTM-based language decoder. This architecture is the same as the work of Anderson et al.~\cite{Anderson_2018_CVPR}.

The LSTM operations are denoted as:
\( \mathbf{h}_{t} = LSTM(\mathbf{x}_t, \mathbf{h}_{t-1})\),  
where, $\mathbf{x}_t$ is the LSTM input vector and $\mathbf{h}_t$ is the LSTM output vector. 
Let the LSTM states for the attention layer and the decoder layer at time step $t$ be $\mathbf{h}^1_{t}$ and $\mathbf{h}^2_{t}$. 
At each time step $t$, the attention LSTM captures contextual information $\mathbf{x}_{t}^1$ by concatenating the previous hidden state of the decoder LSTM, the mean-pooled VSG features ($\bar{\mathbf{f}}=\frac{1}{k}\sum_{i}{\mathbf{f}_i}$)
and the previously generated word representation as following: 
\begin{equation}
\mathbf{x}_{t}^1 = \mathrm{concat}(\mathbf{h}_{(t-1)}^2, \bar{\mathbf{f}}_t, \mathbf{W}_e \mathbf{u}_t)
\vspace{-2mm}  
\end{equation}
where $\mathbf{W}_e$ is the word embedding matrix for vocabulary $\Sigma$ and $\mathbf{u}_t$ is its one-hot encoding at $t$. 

We generate normalized attention weights $\alpha_t$ for the VSG features at time step $t$ as follows: 
\begin{equation}
\begin{split}
\mathbf{a}_{i,t} &= \mathbf{w}^T_{a}\tanh(\mathbf{W}_{fa}\mathbf{f}_i + \mathbf{W}_{ha}\mathbf{h}^1_t) \\
\alpha_{i,t} &= \mathrm{softmax}(\textbf{a}_{i,t})
\end{split}
\vspace{-4mm}  
\end{equation}
where $\mathbf{w}^T_{a} \in \mathbb{R}^H$, $\mathbf{W}_{fa} \in \mathbb{R}^{H\times D_f}$, $\mathbf{W}_{ha} \in \mathbb{R}^{H\times H}$ are learnable weights.
%
The attended VSG features, $\hat{\textbf{f}}_t$, input to the decoder LSTM is thus a convex combination of the input features $\textbf{f}_i$. 
\begin{equation}
\hat{\mathbf{f}}_t = \sum_{i=1}^{N_{f}} \alpha_{i,t}\,\mathbf{f}_i
\vspace{-2mm}  
\end{equation}
The $\mathcal{X}$ formulation allows for learning attention coefficients for each object-level node of the scene graph.

The input to the decoder LSTM consists of the previous hidden state from the attention LSTM layer, and the attention weighted VSG features: 
\(x_{t}^2 = [\textbf{h}_t^1, \hat{\textbf{f}_t}]\).
For a sequence of words $w_1, w_2, ..., w_T$, denoted as $w_{1:T}$, the conditional distribution over possible output words at $t$ is given by: 
$p(w_t |w_{1:t-1}) = \mathrm{softmax}(\mathbf{W}_p \mathbf{h}_t^2 + \mathbf{b}_p)$, 
where, $\mathbf{W}_p \in \mathbb{R}^{|\Sigma|\times H}$ and $\textbf{b}_p \in \mathbb{R}^{|\Sigma|}$ are learned weights and biases. The distribution over complete output sequences is calculated as the product of conditional distributions.


Given a target ground truth captions $w^*_{1:T}$, and a captioning model with parameters $\theta$, we train our encoder-decoder model using one of the two loss functions: \\
(i) Minimize a cross-entropy loss:    
\begin{equation}
\vspace{-4mm}  
L_{XE}(\theta) = - \sum_{t=1}^{T}\log (p_{\theta}(w^*_t | w^*_{1:t-1})) 
\end{equation}
(ii) Maximize a reinforcement learning (RL)-based reward ~\cite{rennie_self-critical_2017}
\begin{equation}
R_{RL}(\theta) = {\mathbb{E}}_{w_{1:T}\overset{}{\sim}p}[rw(w_{1:T})]    
\end{equation}

\par
\noindent
where, $rw(\cdot)$ is the score function (\textit{e.g.\ } the CIDEr metric). 

\section{Experiments}
\label{sec:results}
\subsection{Datasets}
\textbf{The COCO-Captions}~\cite{Chen2015MicrosoftCC}: We conducted the experiments and evaluation our proposed \textbf{SG2Caps} model on the Karpathy split of COCO-Caption dataset for the offline test. This split has $113,287/5,000/5,000$ train/val/test images, each of which has 5 captions. 

\textbf{Visual Genome} \label{VG150} \textbf{(VG)}~\cite{visual_genome16}: We used the VG150 dataset to pretrain a scene graph generator and attribute classifier. VG comes with scene graph annotations, such as 
object categories, object attributes, and pairwise relationships. These are utilized to train an object proposal detector, attribute classifier, and relationship classifier as our VSG parser. 
We first pre-trained a faster R-CNN based object detector ~\cite{tang2020sggcode} using object annotations from Visual Genome with an additional 2-layer multi-layer-perceptron layer for 203 one-vs-all attribute classifiers. Then the Neural Motif ~\cite{neural_motifs_2017} model serves as the ROI head to predict the pair-wise relationships. 

\textbf{Verbs in COCO}: 
The V-COCO dataset contains a subset of COCO images, and was created for evaluating the HOI task. It has annotations for 16K people-instances in 10K images with their actions and associate objects labeled.
We utilize VSGNet ~\cite{VSGNet_2020_CVPR}, a pre-trained HOI model, to generate the inference on COCO images.  

\subsection{Experimental settings} 
\textbf{Processing pseudolabels: }
The VSGs generated by an inference on COCO-caption images utilize the pre-trained MotifNet ~\cite{neural_motifs_2017} as a black box (refer to Section \ref{VG150}). 
The predicted scene graphs are noisy, containing many object proposals many of which are duplicates. They also contain predicted attributes for each proposal and pairwise relationship for all objects. We adapt these VSGs to make them better suited for caption generation. A blackbox VSG generators were trained to achieve high retrieval performance measured by recall. This metric only cares about higher fraction of correct matches to be returned, but does not penalize duplicate objects. However, a VSG to be used for caption generation needs to be free from duplicate objects. 
\par
 We thus discard the less accurate (confidence score below 0.25) object predictions, and apply non-maximum-suppression (NMS) on object proposals 
with intersection-over-union (IOU) threshold of $30\%$. 
We also discard weak relationships, if the confidence score is below 30\% and keep only the best attribute per node provided the confidence is above 90\%. 
\par\textbf{Processing HOI graph:}
HOI graphs are constructed by extracting relationships and attributes from the HOI inference network \cite{VSGNet_2020_CVPR}, which utilizes the instance detection results from detectron2. Images, where \emph{human} objects are detected with a score of 0.5 or more, are selected as inputs to the HOI network. Outputs of the HOI network are $<agent(human) - instrument - object>$ triplets associated with a \emph{HOI relation}, $hr$. 
For example, in the case of `a person hitting a ball with a bat', the triplet takes the form of $<person(agent) - bat - ball>$ for the action `hitting'. To transform such a triplet into a graph, all agents, objects and instruments are considered as the nodes of the graph, and the associated HOI-relation ($hr$) is added as a relationship between the subject and the object. The list of such HOI relations consists of 10 different semantic verb actions. On the other hand, for inferences without $objects$ \textit{e.g.} \emph{stand}, we add them as an attribute \textit{e.g.} \emph{standing} to the $subject$, since it is not possible to form any directed edge between a $subject$ and an $object$. The list of such $object$-less actions that are utilized as attributes consists of 16 different semantic verbs. HOI graphs generated in the above way have such relations or attributes for $47726$ training images and $1912$ test images. Other images transform to graphs with only the detected objects as nodes with only one relation $<object1, AND, object2>$ per image.
Since the goal of utilizing scene graphs for captioning is to enrich the model with objects relationships, we argue that spatial locations of nodes should be leveraged. Our VSG thus contains a list of nodes each consisting of object class label, bounding box, attributes and a set of edges. 
A few object categories (15 such categories) such as \emph{potted plant} from the generated HOI graph don't appear in the captioning vocabulary. In such cases, the HOI node category is mapped to the closest word \emph{plant} from the caption vocabulary. For our experiments, we use the union of pesudolabels and HOI graphs as the VSG of an image. For the inference in language generation, we use greedy search. 

\par
\textbf{Visual Features}
When using additional visual features, we utilize low-dimensional global image features in contrast to the recent captioning methods which use high-dimensional region-level image features (256-D vs 2048-D). We append a projected summary node to the GCN output for each image, where the projected summary node is generated via a learnable projection head applied to the 256-D global pooled F6 features per image. 

\subsection{Results}

Our implementation is built upon the source code ~\cite{sgae_code} of SGAE  ~\cite{SC_autoencoding_Yang_2019_CVPR}. 
Our models are trained on a single NVIDIA 1080Ti GPU running pytorch 0.4.0 in python 2.7.15. 
We evaluate our caption generation model on standard metrics such as BLEU@1(B@1), BLEU@4 (B@4), ROGUe(R), METEOR(M), CIDEr(C) and SPICE(S).

\begin{table}[tb]
\caption{\textbf{Incompatibility between TSG and VSG}. Performance of the caption model trained
on textual scene graph (TSG) while evaluating on different scene graphs as input. TSG row denotes sentence scene graph as inputs. 
$\dagger$ denotes our reproduced results with cross-entropy loss using \cite{sgae_code}. 
PL denotes pseudolabel and (PL + HOI) denotes the union of pseudolabel and HOI graph as the input respectively. 
Inference on TSG-trained model fail to generate decent captions with VSG input. 
}
\centering
\begin{tabular}{l|c|c|c|c|c|c} 
\toprule
Model & B@1 & B@4 & M & R & C & S \\ 
\midrule
TSG $\dagger$ & \textbf{96.2} & \textbf{56.7}  & \textbf{35.5} & \textbf{68.7} & \textbf{158.6} & \textbf{31.2} \\
\midrule
PL & 46.7 & 7.7  & 14.6 & 35.7 & 31.5 & 9.0 \\ 
\textit{PL + HOI} & 50.7 & 9.1  & 15.8 & 37.4 & 38.6 & 9.4 \\
\bottomrule
\end{tabular}
\label{table:TSG-VSG-incompatibility}
\vspace{-5mm}
\end{table}

\par
\textbf{TSG-VSG incompatibility:}
First we show the incompatibility between TSG and VSG in Table \ref{table:TSG-VSG-incompatibility}.
All the entries in this table are from our reproduced models. 
TSG, when used as an input, 
generates excellent captions using the GCN and LSTM language model, as can be seen from the first row. 
The above TSG-caption model was trained with cross-entropy loss.  
However, if we simply use that model and perform inference using 
the pseudolabel 
as inputs, it performs significantly worse.
Although the HOI augmentation improves the CIDEr score by 7 points, the overall caption generation performance still remains poor.
The takeaway message is that although VSG and TSG are similar type of representations, but they are not directly compatible to each other for caption generation. 


\par
\textbf{Caption generation performance comparison:}
In our work, we focus on scene graphs for caption generation, and thus limit our experiments and comparisons to the LSTM language model. Table \ref{table:caption_generation_results} shows the main result of our proposed \textbf{SG2Caps} method. 
The bottom half of the table denotes methods that use only scene graph labels as input to the captioning model. 
Our graph construction differs from ~\cite{lanterns19} in a few ways. Wang \etal used FactorizableNet \cite{FactorizableNet_Li_2018_ECCV} as the relation detection model on top of RPN from the Faster R-CNN. Wang \etal didn't use 
bounding box locations which we find important for caption generation. We use a simpler relation model \cite{neural_motifs_2017} for scene graph detection. 
\textbf{SG2Caps} also outperforms the most recent captioning model, Sub-GC ~\cite{zhong2020comprehensive} adapted for \emph{G-only} mode. Since the paper does not report the \emph{G-only} model performance, we have adapted their code ~\cite{subGC_code} so as to use only the scene graphs (excluding the visual features) for the captioning. 
Please note, we used their scene graphs ~\cite{subGC_code} while reproducing their \emph{G-only} captioning model which has 1600 object classes while our object class vocabulary contains only 230 words. The graph construction and the GCN architecture in  ~\cite{zhong2020comprehensive} both differ from SG2Caps. \textbf{SG2Caps} is able to process variable number of nodes per VSG, while \cite{zhong2020comprehensive} can operate on a fixed number of nodes per VSG per the architecture design.
Our SG2Caps significantly outperforms the existing \emph{G only} models by large margin. 
With HOI augmentation and bounding box feature, SG2Caps produces competitive results 
close to the  SoTA models that rely on object detection CNN features.  

\begin{table}[tb]
\centering
\caption{\textbf{Comparison with State-of-the-art captioning models}. We compare with both image captioning SoTA methods (that use both region-level visual features and scene-graphs labels) and methods that use scene graphs as only input. 
Here, all methods use LSTM language model. 
RSG-G1 and RSG-G2 refer to without and with gate-wise gating in the G-only setup from RSG. All results are from the corresponding papers.
$\dagger$ denotes our implementation based on \cite{subGC_code}.
$\oplus$ denotes captioning models that use global visual features. SG2Caps - RL denotes CIDEr-based optimization, all others use cross-entropy loss.
}
\resizebox{1.0\linewidth}{!} 
{ 
\begin{tabular}{l|c|c|c|c|c} 
\toprule
Model & B@4 & M & R & C & S \\ 
\midrule
\multicolumn{6}{c}{\textit{visual scene graph + visual features}}\\ \midrule
SGAE \cite{SC_autoencoding_Yang_2019_CVPR} &  36.9 & 27.7 & 57.2 & 116.7 & 20.9\\
R-SCAN \cite{visrelprior_2019} &  - & - & - & 114.9 & 20.9 \\ 
SubGC \cite{zhong2020comprehensive} &  36.9 & 27.9 & 56.8 & 114.8 & 20.8 \\
KMSL \cite{tmm19}  & 33.8 & - & 54.9 & 110.3 & 19.8 \\
Attribute \cite{Wu2018ImageCA}  &31.0 & 26.0 & - & 94.0 & - \\
SGC \cite{AACL-IJCNLP2020}  & 35.5 & - & 56.0 & 109.9 & 19.8 \\
RSG \cite{lanterns19} & 34.5 & 26.8 & 55.9 & 108.6 & 20.3 \\
SCST:Att2in \cite{rennie_self-critical_2017}$\oplus$ & 31.1 & 26.0 & 54.3 & 101.3 & - \\
\textbf{SG2Caps (ours)}$\oplus$ &  32.6 & 26.4 & 55.0 & 106.6 &  19.8 \\
\midrule
\multicolumn{6}{c}{\textit{visual scene graph only}}\\ 
\midrule
RSG-G1 \cite{lanterns19}  & 22.8 & 20.6 & 46.7 & 66.3 & 13.5 \\
RSG-G2 \cite{lanterns19}   & 22.9 & 21.1 & 47.5 & 70.7 & 14.0 \\
SubGC \cite{zhong2020comprehensive}$\dagger$ & 31.5 & 25.5 & 53.4 & 98.9 & 18.4 \\
\textbf{SG2Caps (ours)} & \textbf{32.0} & \textbf{26.2} & \textbf{54.9} & \textbf{104.4} &  \textbf{19.5} \\
\textbf{SG2Caps - RL (ours)} & \textbf{33.0}  & \textbf{26.2} & \textbf{55.6} & \textbf{112.3} & \textbf{19.4} \\
\bottomrule
\end{tabular}
? }
\label{table:caption_generation_results}
\vspace{-7mm}
\end{table}

We also compare our model with the state-of-the-art captioning models that use both the visual features and explicit scene graphs.
All such methods utilized 2048-D object detection CNN features for each salient region.
Wu \etal \cite{Wu2018ImageCA} utilized classified object attributes on top of the visual features. 
The ``Know more, say less'' model (KMSL) extracts features for objects and relations based on the scene graph, which are passed through two attention heads and finally combined using a flat attention head. 
R-SCAN~\cite{visrelprior_2019} utilized soft alignment using attention and visual features to associate the region features and the relation features to the word features. Their scene graph generation was trained on Vrr-VG dataset~\cite{VrR-VG} instead of Visual Genome dataset widely used in other papers.
RSG ~\cite{lanterns19} explored on how to integrate relation-aware scene graph features encoded by Graph Convolution with region-level image features to boost image
captioning performance. SGC~\cite{AACL-IJCNLP2020} employed conditional graph-attention network to fuse scene graph features and region-level visual features and performed in-depth study on when scene graphs are useful for image captioning and when they are not.
Sub-GC~\cite{zhong2020comprehensive}, although focused on sub-graph based comprehensive and diverse caption generation, they also report competitive image captioning results based on the full-graph. We include it as one of the most recent captioning model for comparison. SGAE \cite{SC_autoencoding_Yang_2019_CVPR} reported the best caption generation performance on our test dataset. SGAE utilized a shared memory for leveraging language inductive bias from the textual scene graphs and employed spatial graph convolution on scene graph representation and region-level visual features. These methods apply graph convolutions on high dimensional visual features, increasing the memory footprint of the captioning model.
Unlike high dimensional (2048-D) region-level visual features, we utilize low-dimensional (256-D) image-level visual features in our visual-feature augmented SG2Caps framework. For example, if an image has 20 objects on an average, SG2Caps needs to operate on 160$\times$ fewer dimensional inputs per image, resulting into significantly lower memory footprint while operating on a batch of 100 images. 
Augmenting the global visual features in the SG2Caps framework results in additional 2-point gain of CIDEr score and overall performs competitively with best captioning models requiring graph convolutions on high-dimensional visual features. 
Please note that our aim is to avoid the GCN operations on the high-dimensional feature space and keep the memory footprint low with a negligible parameter overhead, and thus we avoid the use of object detection CNN features. We compared with the best captioning model that uses global feature~\cite{rennie_self-critical_2017}. However, SG2Caps can also be extended to use the object detection CNN features per VSG node similar to existing state-of-the-art methods.
\begin{figure*}[t] 
\centering
    \includegraphics[width=0.92\linewidth, height=10cm,keepaspectratio]{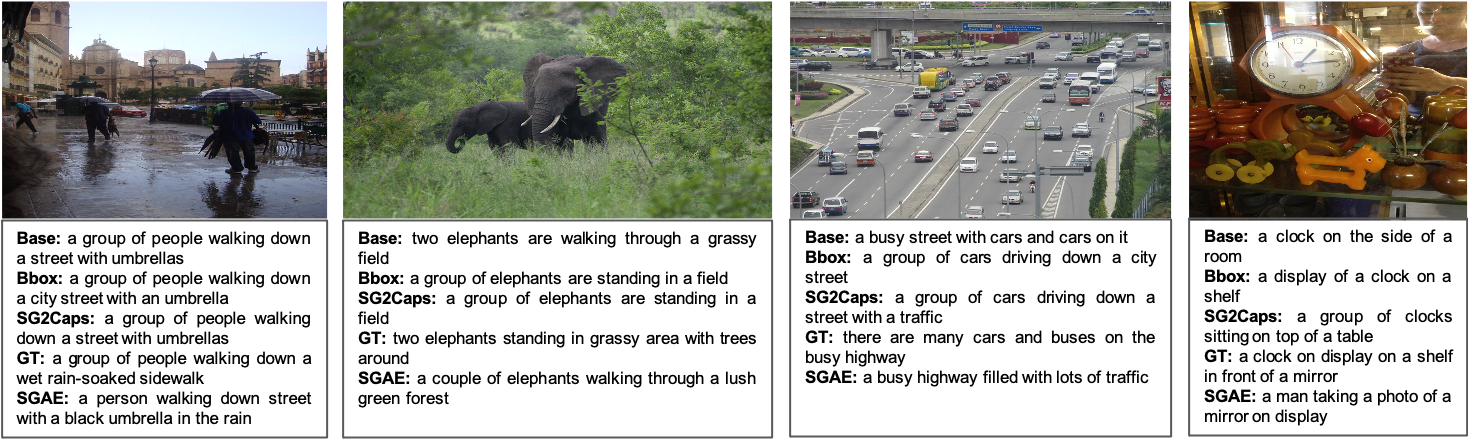}
\caption{Qualitative examples: generated captions from different baseline models from Karpathy test split for COCO image ids \#177861, \#45710, \#553879, \#87399 respectively.} 
\label{fig:subjective_results}
\vspace{-5mm}
\end{figure*}
Results in Table \ref{table:caption_generation_results} show that our SG2Caps substantially minimizes the gap between the performances of state-of-the-art caption generation models and a \emph{scene-graph-only} model.
We leverage visual features in a memory-efficient way. The visual-feature augmented SG2Caps is also able to retain the low-memory footprint.  
We note that the number of trainable parameters for captioning models that use both the high dimensional CNN features and scene graph labels such as SGAE are significantly higher than our scene graph only model. For example, SG2Caps has $49\%$ fewer parameters comparing with SGAE~\cite{SC_autoencoding_Yang_2019_CVPR} full-model (21M vs 41M) for the same language decoder model. The optional visual feature fusion model has an overhead of 0.03M parameters.
\begin{table}[tb]
\caption{
\textbf{Ablation results:} Performance of our \textbf{SG2Caps} model, when the object groundings and HOI are leveraged. All models were trained with Cross-entropy loss. 
}
\resizebox{1.0\linewidth}{!} 
{ 
\begin{tabular}{l|c|c|c|c|c|c|c|c|c|c} 
\toprule
Model & bbox & HOI & g\_feat & B@1 & B@4 & M & R & C & S \\ 
\midrule
Baseline & \xmark & \xmark & \xmark & 72.0 & 29.3  & 24.5 & 52.6 & 92.6 & 17.8 \\
BBox  & \checkmark & \xmark & \xmark & 73.1 & 30.3  & 25.1 & 53.4 & 97.4 & 18.4  \\
Global & \xmark & \xmark & \checkmark & 73.3 & 30.7 & 25.5 & 53.9 & 98.1 &  18.5 \\
SG2Caps & \checkmark & \checkmark & \xmark & 75.0 & 32.0 & 26.2 & 54.9 & 104.4 &  19.5 \\ 
SG2Caps-G & \checkmark & \checkmark & \checkmark & \textbf{75.0} & \textbf{32.6} & \textbf{26.4} & \textbf{55.0} & \textbf{106.6} &  \textbf{19.8} \\ 
\bottomrule
\end{tabular}
}
\label{table:ablation_results}
\vspace{-5mm}
\end{table}
%
\par
\textbf{Ablation experiments:}
Table \ref{table:ablation_results} shows the results of our ablation experiments. We observe that when the post-processed pseudolabels are directly used for training a captioning model, it can already produce decent captions. This is our baseline model. 
The BBox model corresponds to the node groundings on top of the baseline. Note that when we incorporate the node groundings in the pseudolabels, our BBox model achieves a 5-point gain in CIDEr score. SG2Caps, our final model, uses HOI graph atop BBox model. This HOI augmentation results in further 7 point gain in CIDEr score. An extremely light-weight global feature fusion module boosts the SG2Caps model by further 2 points in CIDEr score.
\par
\begin{figure} 
\centering
    \includegraphics[width=0.9\linewidth]{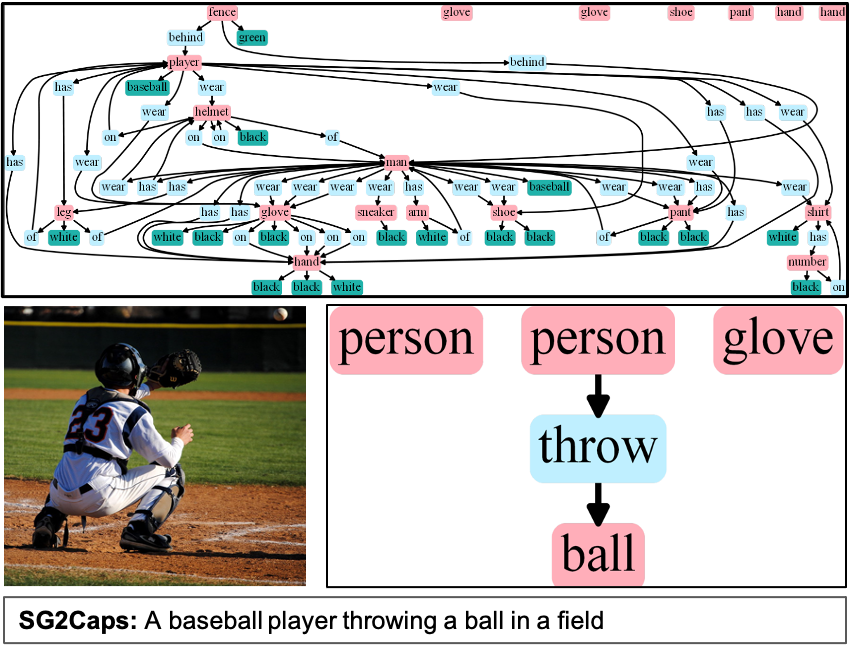}
    \includegraphics[width=0.9\linewidth]{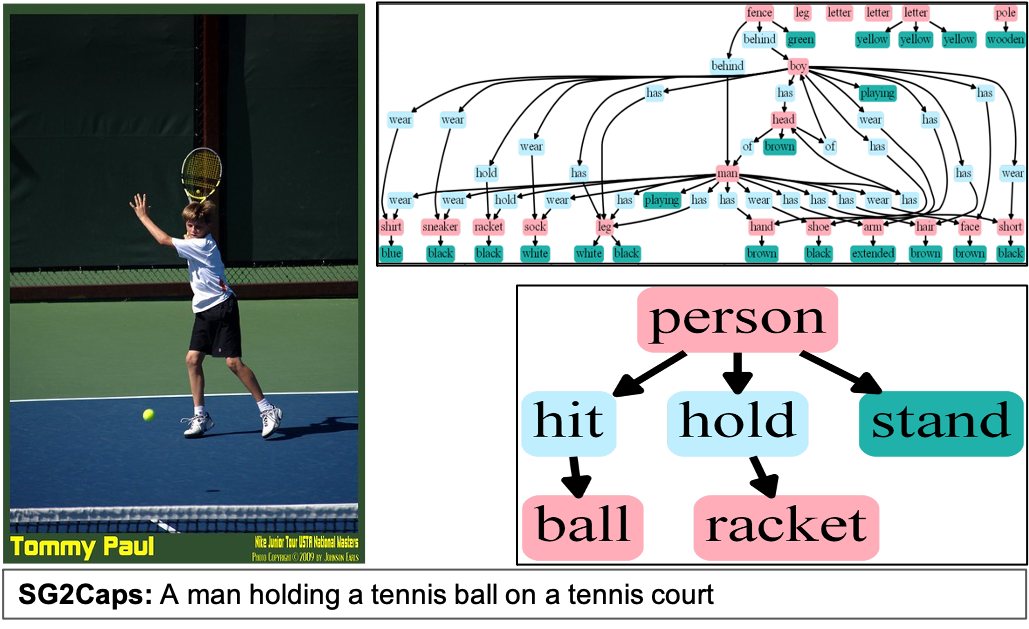}
    \caption{Example VSGs consisting of noisy pseudolabel, HOI graphs and corresponding captions output by SG2Caps.}
\label{fig:subjective2}
\vspace{-5mm}
\end{figure}

\textbf{Qualitative results:}
Our baseline generates convincing caption sentences. Fig.~\ref{fig:subjective_results} shows that the caption quality improves over the baseline to BBox and SG2Caps models. We also show captions from SGAE~\cite{SC_autoencoding_Yang_2019_CVPR}, one of the best captioning models that uses both visual features and scene graphs. SG2Caps is able to generate high-quality and competitive captions without using visual features (see Fig.~\ref{fig:subjective_results}). %
Fig.~\ref{fig:subjective2} shows noisy pseudolabels and HOI graphs and their corresponding SG2Caps output for example images (image\_ids 548361 and 258628) from COCO Karpathy test split. The HOI graph detects \emph{semantic verb} relations which are complementary to the pseudolabel that detects many other objects, relations and their attributes. As seen in the first example, the words in the output caption come from both the pseudolabels (such as \emph{baseball player}, \emph{ball}, \emph{field}) and the HOI-graph (\emph{throw}, \emph{ball}).   
%

\section{Conclusion}
\label{sec:conclusion}
Explicit encoding of objects, attributes and relations are useful information for image captioning. However, blindly using visual scene graphs for captioning fails to produce reasonable caption sentences. The proposed SG2Caps pipeline enables networks pre-trained for (1) SGDet on other scene graph datasets, and (2) semantic roles on HOI datasets to greatly reduce the gap in accuracy on COCO caption datasets – indicating strong captioning models can be achieved with low dimensional objects and relations label space only. These results further strengthen our defense of scene graph for image captioning. We hope our observations can open up new opportunities for vision and language research in general.


\section*{Supplementary Material}
\label{sec:supplementary}

\section*{SG2Caps with global visual features:}
For the global features, we first extract P6 features from the ResNeXt101-FPN backbone using global max pooling, resulting in 256-D global visual features per image. This is treated as the global visual features - one set of features for each image. Next, we employ a global feature fusion strategy as follows.
We add a summary projection node in the output graph embedding. The summary node is generated using global pooling of the image-level F6 features, followed by a projection layer that takes the 256-dimensional summary node feature to 128-dimensional vector, and then a ReLU non-linearity. The projected summary node dimension is compatible with the GCN embedding size. This projected summary node feature is concatenated with the scene graph embedding vector generated in the SG2Caps. 
This feature fusion has a negligible parameter overhead (256$\times$128 additional parameters) on top of SG2Caps. Using scene graph embeddings results in 7 point gain in CIDEr score (106.6 vs 99.5).
It also means an extremely light-weight global feature fusion module boosts the SG2Caps model by further 2 points in CIDEr score (106.6 vs 104.4) when global visual features are being used.  



\section*{Training Details}
The main model, SG2Caps was trained on a single Tesla M40 24GB. The batch size was set at 300, with an accumulate number of 1. The learning rate was initially set at $5\mathrm{e}{-4}$, with a learning rate decay of .8 that applied every 3 epochs. This model was trained for 8.5k iterations using the cross entropy loss before moving to the reinforcement learning, which was trained for an additional 3k iterations. The SG2Caps-G model was trained with two Tesla M40 24GBs, with a batch size of 200 with an accumulate number of 2. The rest of the training details are the same. This model trained on the cross entropy loss for 8.5k iterations.

\begin{figure*}[t] 
\centering
    \includegraphics[width=0.92\linewidth, height=10cm,keepaspectratio]{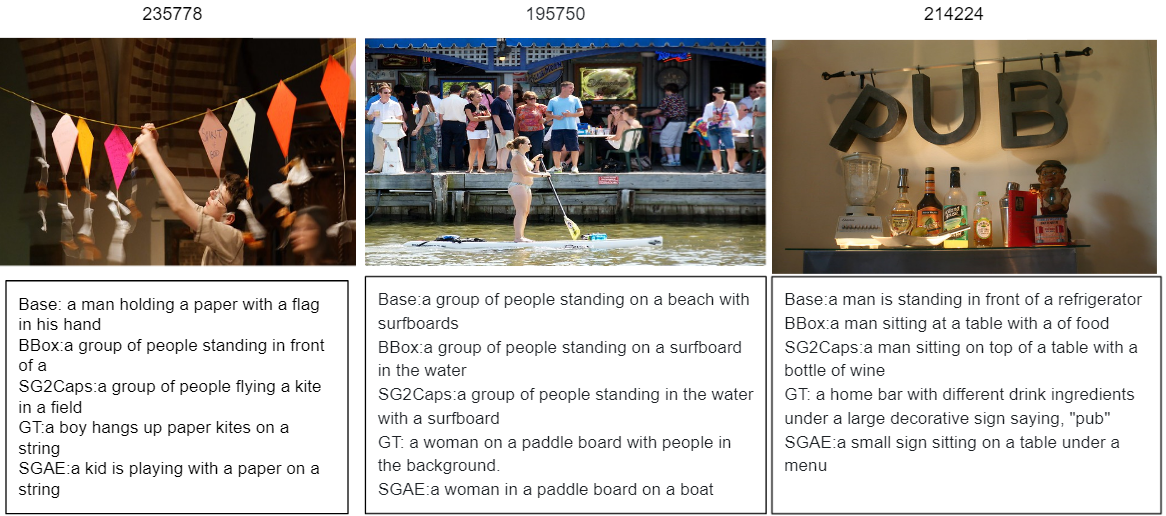}
\caption{Qualitative examples where SG2Caps  performed worse than SGAE: generated captions from different baseline models from Karpathy test split for COCO image ids \#235778, \#195750, \#214224 respectively.} 
\label{fig:bad_results}
\vspace{-5mm}
\end{figure*}
\begin{figure*}[t] 
\centering
    \includegraphics[width=0.92\linewidth, height=10cm,keepaspectratio]{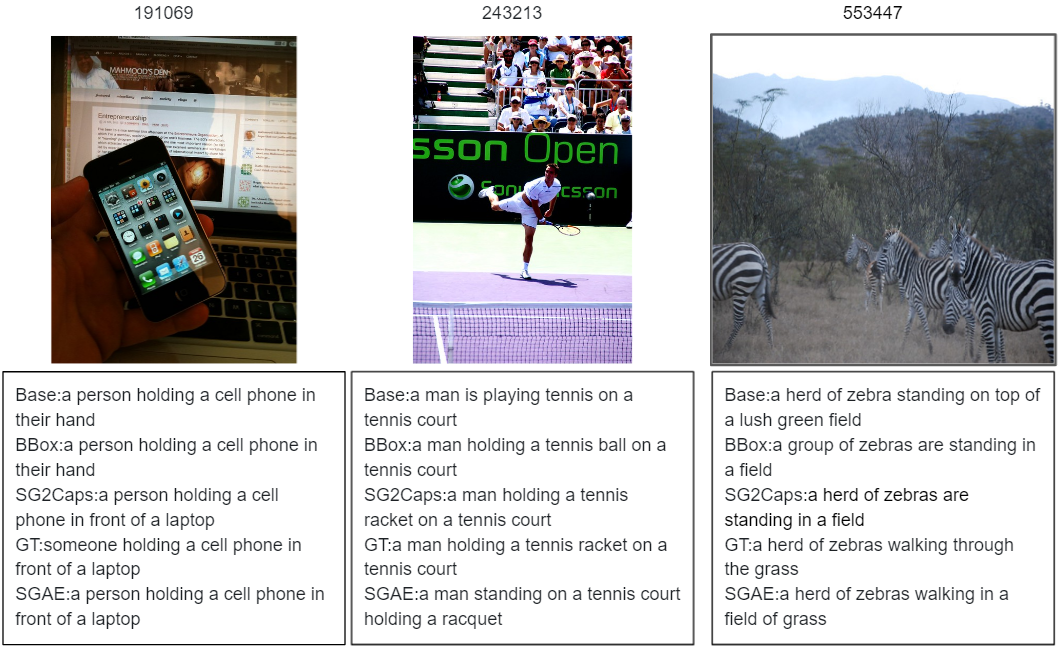}
\caption{Qualitative examples where SG2Caps is as good as SGAE: generated captions from different baseline models from Karpathy test split for COCO image ids \#191069, \#243213, \#553447 respectively.} 
\label{fig:ok_results}
\vspace{-5mm}
\end{figure*}
\begin{figure*}[t] 
\centering
    \includegraphics[width=0.92\linewidth, height=10cm,keepaspectratio]{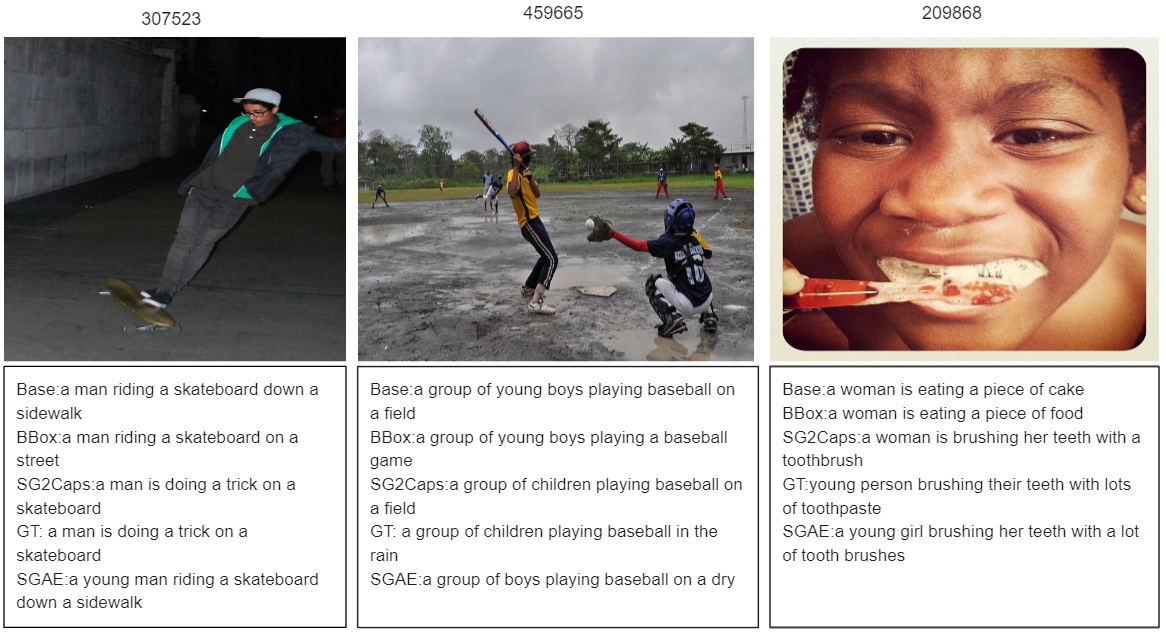}
\caption{Qualitative examples where SG2Caps is arguably better than SGAE: generated captions from different baseline models from Karpathy test split for COCO image ids \#307523, \#459665, \#209868 respectively.} 
\label{fig:good_results}
\vspace{-5mm}
\end{figure*}

\section*{Additional Examples}

We have anecdotally compared the SG2Caps results with the SGAE for several images.
Results in Figure \ref{fig:ok_results} show three such images where our model performed closely to the SGAE model. This is good because it shows that our SG2Caps model can produce similar captions in comparison to SGAE, even though our model has almost half of the trainable parameters of SGAE.
Results in Figure \ref{fig:good_results} show three such images where SG2Caps  generated better captions than the SGAE model.

Results in Figure \ref{fig:bad_results} show different images where our SG2Caps model performed worse, in comparison to SGAE. 
As is seen in Figure \ref{fig:bad_results} in image \#195750, where the model sees multiple humans, SG2Caps returns a caption stating that a group of people doing an action, rather than one or two people doing the action with the rest of the humans just being part of the crowd.

{\small
\bibliographystyle{ieee_fullname}
\bibliography{main}
}

\end{document}